\documentclass{article}
\usepackage{ijcai20}

\usepackage{url}
\usepackage[hidelinks]{hyperref}
\usepackage[utf8]{inputenc}
\usepackage[small]{caption}
\usepackage{graphicx}
\usepackage{amsmath}
\usepackage{booktabs}
\usepackage[american]{babel}
\usepackage[T1]{fontenc}
\usepackage{balance}

\makeatletter
\renewcommand*{\p@section}{\S\,}
\renewcommand*{\p@subsection}{\S\,}
\makeatother

\newcommand{\erhao}{\fontsize{21pt}{\baselineskip}\selectfont}

\usepackage{times}
\usepackage{bm}

\usepackage{epsfig}
\usepackage{graphicx}
\usepackage[caption=false]{subfig}

\usepackage{pifont}
\usepackage{bbding}

\usepackage{amsmath}
\usepackage{amssymb}
\usepackage[american]{babel}
\usepackage{arydshln}
\usepackage{enumitem}
\usepackage[normalem]{ulem}

\usepackage{array}
\usepackage{booktabs} %
\usepackage{multirow} 

\usepackage{xspace}
\usepackage[table]{xcolor}
\usepackage{setspace}

\definecolor{dgreen}{rgb}{0.04, 0.565, 0} 
\definecolor{cyan}{rgb}{0, 0.5, 0.6} 
\definecolor{Darkviolet}{rgb}{0.58, 0, 0.83} 

\definecolor{LightCyan}{rgb}{0.95,0.95,0.95}

\definecolor{lr}{rgb}{0.8275, 0.9490, 0.7490}
\definecolor{ir}{rgb}{0.9961, 0.8078, 0.8078}
\definecolor{sk}{rgb}{0.9961, 0.9088, 0.5508}
\definecolor{tx}{rgb}{0.7922, 0.8588, 0.9843}

\newcommand{\ie}{{\it i.e.}}
\newcommand{\eg}{{\it e.g.}}
\newcommand{\vs}{{\it vs.}~}
\newcommand{\etal}{{\it et al.}}

\newcommand{\heading}[1]{\noindent\textbf{#1}}

\newcommand{\figref}[1]{Figure~\ref{fig:#1}}%
\newcommand{\tabref}[1]{Table~\ref{tab:#1}} %

\newcommand{\tb}[1]{\textbf{#1}}

\newcommand{\ignore}[1]{}   %

\newcolumntype{L}[1]{>{\raggedright\arraybackslash}p{#1}}
\newcolumntype{C}[1]{>{\centering\arraybackslash}p{#1}}
\newcolumntype{R}[1]{>{\raggedleft\arraybackslash}p{#1}}

\newcommand{\circledlr}[1]{\fcolorbox{black}{lr}{\small{#1}}}
\newcommand{\circledir}[1]{\fcolorbox{black}{ir}{\small{#1}}}
\newcommand{\circledtx}[1]{\fcolorbox{black}{tx}{\small{#1}}}
\newcommand{\circledsk}[1]{\fcolorbox{black}{sk}{\small{#1}}}

\usepackage{xcolor}
\colorlet{dark-blue}{blue!50!black}
\colorlet{dark-cyan}{cyan!75!black}
\colorlet{dark-purple}{purple!50!black}
\colorlet{dark-red}{red!75!black}
\colorlet{dark-green}{green!75!black}
\colorlet{dark-orange}{orange!50!black}
\colorlet{dark-gray}{black!75}
\colorlet{light-gray}{black!30}
\colorlet{hidden}{light-gray}
\colorlet{todo}{red!85!black}
\colorlet{todoref}{purple!70!black}
\definecolor{nice-red}{HTML}{E41A1C}
\definecolor{nice-orange}{HTML}{FF7F00}
\definecolor{nice-yellow}{HTML}{FFC020}
\definecolor{nice-green}{HTML}{4DAF4A}
\definecolor{nice-blue}{HTML}{377EB8}
\definecolor{nice-purple}{HTML}{984EA3}

\setlist[itemize]{leftmargin=*}

\makeatletter
\def\@fnsymbol#1{\ensuremath{\ifcase#1\or \dagger\or \ddagger\or
\mathsection\or \mathparagraph\or \|\or **\or \dagger\dagger
\or \ddagger\ddagger \else\@ctrerr\fi}}

\newcommand{\printfnsymbol}[1]{%
  \textsuperscript{\@fnsymbol{#1}}%
}
\makeatother

\hypersetup{
    colorlinks=true,%
    citecolor=dark-blue,%
    filecolor=dark-blue,%
    linkcolor=dark-blue,%
    urlcolor=magenta
}

\newcommand{\datatag}[2]{\setlength{\fboxsep}{1pt}\colorbox{#1}{{#2\hspace{1pt}}}}

\title{Beyond Intra-modality: A Survey of Heterogeneous Person Re-identification}

\author{
Zheng Wang\thanks{Equal contribution}$^1$\and
Zhixiang Wang\printfnsymbol{1}$^{2}$\and
Yinqiang Zheng$^1$\and
Yang Wu$^3$\and \\
Wenjun Zeng$^4$ \And
Shin'ichi~Satoh$^{1,5}$\\
\affiliations
$^1$National Institute of Informatics \and $^2$National Taiwan University\\
$^3$Kyoto University \and
$^4$Microsoft Research Asia \and $^5$The University of Tokyo\\
\emails
\small{\texttt{\{wangz, yqzheng, satoh\}@nii.ac.jp,
\{wangzx1994, wuyang0321\}@gmail.com, wezeng@microsoft.com}}
}

\begin{document}

{\onecolumn

\noindent \textbf{\erhao{Beyond Intra-modality: A Survey of Heterogeneous Person Re-identification}}

\vspace{2cm}

\noindent {\Large{Zheng Wang, Zhixiang Wang, Yinqiang Zheng, Yang Wu, Wenjun Zeng, Shin'ichi Satoh}}

\Large
\vspace{2cm}

\noindent Follow-up updates are available at: \\
\ \ \ \ \ \ \ \ \ \ \ \ \url{https://github.com/lightChaserX/Awesome-Hetero-reID}

\vspace{1cm}

\noindent For reference of this work, please cite:

\vspace{1cm}
\noindent Zheng Wang, Zhixiang Wang, Yinqiang Zheng, Yang Wu, Wenjun Zeng, Shin'ichi Satoh.
``Beyond Intra-modality: A Survey of Heterogeneous Person Re-identification.''
In \emph{IJCAI.} 2020.

\vspace{1cm}

\noindent Bib:\\
\noindent\normalsize
@inproceedings\{wang2020beyond,\\
\ \ \  title=\{Beyond Intra-modality: A Survey of Heterogeneous Person Re-identification\},\\
\ \ \  author=\{Wang, Zheng and Wang, Zhixiang and Zheng, Yinqiang and Wu, Yang and Zeng, Wenjun and Satoh, Shin'ichi\},\\
\ \ \  booktitle=\{IJCAI\},\\
\ \ \  year=\{2020\}\\
\}
}

\twocolumn

\maketitle

\begin{abstract}
An efficient and effective person re-identification (ReID) system relieves the users from painful and boring video watching and accelerates the process of video analysis. Recently, with the explosive demands of practical applications, a lot of research efforts have been dedicated to heterogeneous person re-identification (Hetero-ReID). In this paper, we provide a comprehensive review of state-of-the-art Hetero-ReID methods that address the challenge of inter-modality discrepancies. According to the application scenario, we classify the methods into four categories --- low-resolution, infrared, sketch, and text. We begin with an introduction of ReID, and make a comparison between Homogeneous ReID (Homo-ReID) and Hetero-ReID tasks. Then, we describe and compare existing datasets for performing evaluations, and survey the models that have been widely employed in Hetero-ReID. We also summarize and compare the representative approaches from two perspectives, \ie, the application scenario and the learning pipeline. We conclude by a discussion of some future research directions. \emph{Follow-up updates are avaible at: \url{https://github.com/lightChaserX/Awesome-Hetero-reID}}
\end{abstract}

\section{Introduction}
\label{sec:introduction}

Person re-identification (ReID) is a human-centric AI technology that finds person of interest in a large amount of videos. It facilitates various applications that require painful and boring video watching, including searching for video shots related to an actor of interest from TV series, finding a lost child in a shopping mall from camera videos, and re-identifying suspects in video surveillance systems. Its efficiency and effectiveness accelerate the process of video analysis. In recent years, we have made significant advances in general ReID. The performance is remarkably high on some public datasets~\cite{luo2019strong,wang2019incremental,wang2019sparse,zheng2019pose,zhou2019discriminative}, for example, 96.1\% Rank-1 accuracy on the Market-1501 dataset~\cite{zheng2015scalable}, which even outperforms human~\cite{zhang2017alignedreid}.
However, the general ReID assumes an ideal scenario for brevity where all person images are captured in the daytime with the visible spectrum, and have sufficient details to represent a person. Considering all these images are in the desired modality, we name person ReID under this scenario Homogeneous Person Re-identification (Homo-ReID). It %
just counts the challenges of intra-modality discrepancies, such as pose and viewpoint changes~\cite{zheng2019pose,zhang2019densely}, background and illumination variations~\cite{zeng2019illumination}, and occlusions~\cite{zheng2015partial,huang2015sparsity}.

\begin{figure}[!tb]
\centering
    \includegraphics[width=0.88\columnwidth]{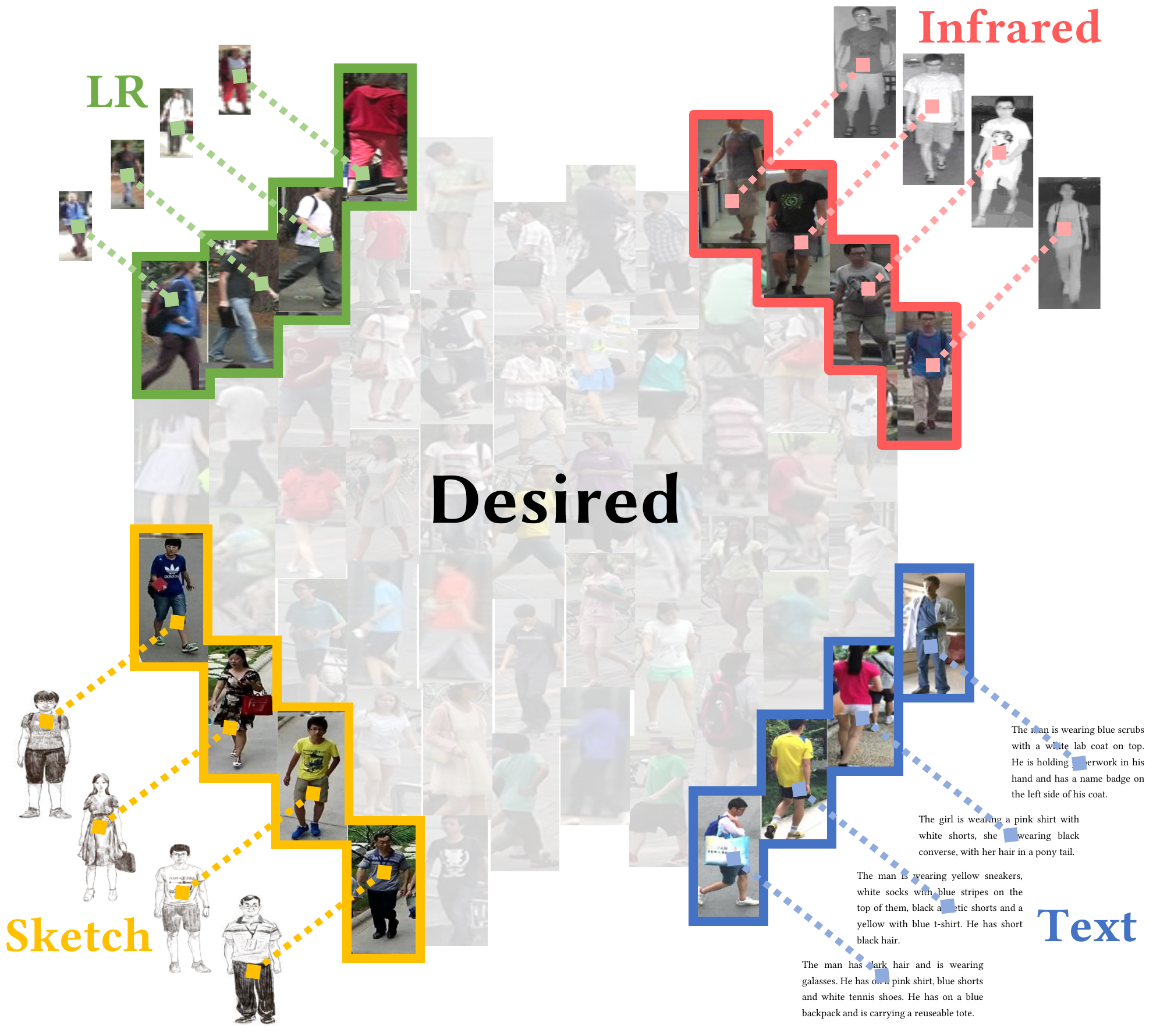}
  \caption{Scope of heterogeneous person re-identification studied in this survey. `LR' indicates low-resolution. The examples are selected respectively from Hetero-ReID LR dataset (MLR-VIPeR), Hetero-ReID IR dataset (SYSU-MM01), Hetero-ReID Sketch dataset (PKU-Sketch) and Hetero-ReID Text dataset (CUHK-PEDES). The dataset details can refer to \tabref{dataset}.}
  \label{fig:problem}
\end{figure}

In real-world applications, however, it is impractical to assume the data source is always in such a desired modality. For example, to find a criminal, the system have to check person images in low-resolution (LR), or captured by infrared (IR) cameras when the illumination condition is not sufficient. Moreover, witnesses' descriptions (in terms of text) and sketches drawn by artists shall also be used as the cues. These scenarios face challenges beyond the intra-modality discrepancy investigated in Homo-ReID as shown in \figref{problem}. Algorithms developed under the desired modality lose their effectiveness when applied to these scenarios.

To improve the practical capability of the ReID system, we should make more efforts to bridge the gap between %
different camera specifications and settings (\eg, low- \vs high-resolution data), different sensory devices (\eg, infrared \vs visible light devices), and reproduction of human memory and direct recording by a camera (\eg, sketch/text description \vs digital images). We define this practical person retrieval Heterogeneous Person Re-identification (Hetero-ReID).

Encouragingly, in the past few years, we have already seen quite a number of remarkable signs of progress on Hetero-ReID, which are primarily assisted by a growing variety of Hetero-ReID benchmark datasets allowing direct comparison of different methodologies. {While Homo-ReID attracts much attention in the research community, we feel that it is the right time to call for more attention and research efforts to the important and still largely unexplored area of Hetero-ReID by providing a quality survey}. This paper offers a comprehensive and up-to-date review of the diverse and growing array of Hetero-ReID techniques.

\heading{The differences between this survey and the previous ones.} 
To our knowledge, there are few reviews in the ReID field. \cite{nambiar2019gait} explored the ReID applications built on gait sequences, which is a special and different focus. \cite{vezzani2013people,zheng2016person,gou2018systematic} focused on Homo-ReID. \cite{vezzani2013people} made a multidimensional overview of Homo-ReID. \cite{zheng2016person} described critical future directions and briefed some important yet under-developed issues. \cite{gou2018systematic} conducted a systematic evaluation with different features and metrics. These works well position the current progress of Homo-ReID. However, they have no investigation on Hetero-ReID and confine to address the intra-modality discrepancy. \cite{leng2019survey,ye2020deep} started to look into some inter-modality challenges, but provided a limited summary of current efforts or problems present in Hetero-ReID. Note that in other areas, face recognition~\cite{ouyang2016survey} and image retrieval~\cite{wang2016comprehensive} also raise the urgent requirements of addressing the inter-modality discrepancy. This survey seeks to provide a summary of current research from different perspectives.

\heading{Goals of this survey.} 
We aim to 
(i) present the idea of Hetero-ReID and raise awareness of this problem;
(ii) thoroughly review the literature on Hetero-ReID and provide a panorama for researchers in other fields, with which they can quickly understand and step into the new area; 
(iii) give comprehensive guidance of future directions to peers in our community.

\heading{Contributions.} 
(i) We highlight the difference between Hetero-ReID and Homo-ReID, and conduct a systematic review of Hetero-ReID beyond intra-modality, where the inter-modality discrepancy is the main challenge. (ii) We categorize related works into different modalities on which they operate and summarize the state-of-the-arts. As \figref{problem} shows, we consider four cross-modality application scenarios: Low-resolution (LR), Infrared (IR), Sketch, and Text. (iii) We analyze representative methods and identify future directions in this research field to share the vision and expand the horizons of Hetero-ReID.

\tabcolsep=4pt
\begin{table}[t]
\renewcommand{\arraystretch}{1.2}
\caption{Comparison between Homo-ReID and Hetero-ReID. `\#Publications' represents the number of related publications. We use CMC-1 values (\%) to show the performances (higher is better). Note that the results are recoded by the end of 2019.}
\label{tab:difference}
\centering
\resizebox{\columnwidth}{!}{
\begin{tabular}{ l c c }
\toprule
                        &\tb{Homo-ReID}                   &\tb{Hetero-ReID}\\
\midrule
\midrule
Media Type              &Desired image                & $+$ \datatag{lr}{LR} / \datatag{ir}{IR}/ \datatag{sk}{Sketch} / \datatag{tx}{Text}\\
Participant             &Machine                     & Machine (+Human)\\
Main Challenge          &Intra-modality              &Intra- {\color{cyan}{$+$ Inter-modality}} \\
\#Publications          &$>$1000                     &$<$100\\
Performance             &96.1                        &\datatag{lr}{42.50} / \datatag{ir}{28.90} / \datatag{sk}{34.00} / \datatag{tx}{53.14} \\
\bottomrule
\end{tabular}}
\end{table}

\begin{figure}[b]
    \centering
    \includegraphics[width=0.85\columnwidth]{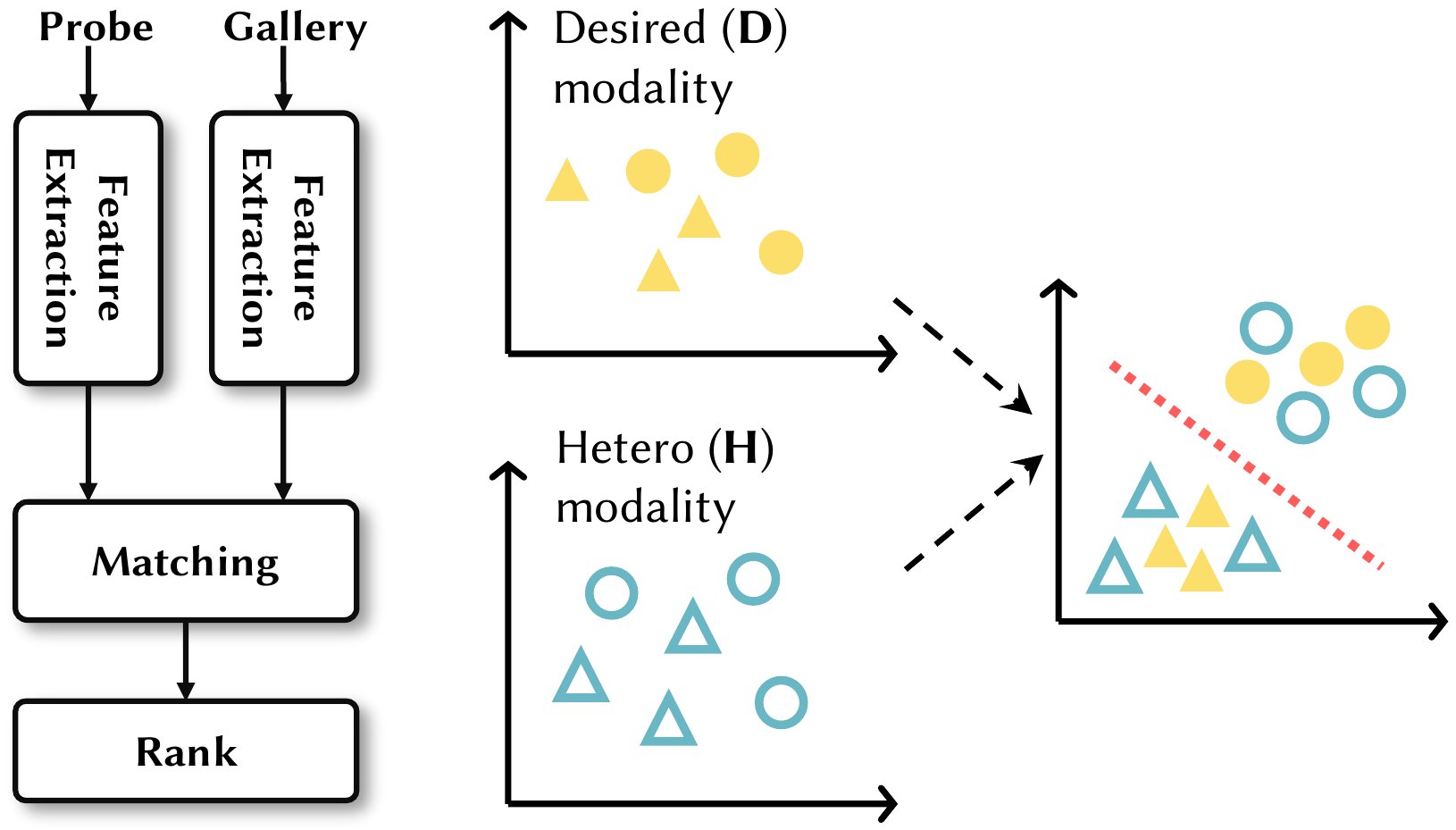}
    \caption{The diagram of Hetero-ReID.}
    \label{fig:diagram}
\end{figure}

\section{Person Re-identification}
\label{sec:he-reid}

\subsection{ReID Diagram} 
ReID system always consists of a feature extraction part and a descriptor matching part. At first, feature vectors of a gallery $\mathcal{G} = \{f_{g_1}, f_{g_2}, ..., f_{g_N}\}$ are extracted by the pre-trained feature extraction part. When we input a probe into the system, the probe's feature vector $f_{p}$ is obtained by the feature extraction part. After that, the feature vector $f_{p}$ matches against feature vectors $\mathcal{G}$ by the descriptor matching part. For Homo-ReID, both probe samples and gallery samples are from the desired modality (daytime, visible spectrum, and high-resolution images). While for Hetero-ReID, the probe sample or gallery samples are from another modality (see \figref{diagram}).

\label{sec:comparison}
\subsection{Hetero-ReID \vs Homo-ReID} Homo-ReID only faces the challenges of persons' appearance changes, and person images are captured in the desired modality with only intra-modality discrepancy. Following the diagram of ReID, researches tried to design robust and discriminative features or learn effective distance metrics. For more details, the readers can refer to Homo-ReID surveys~\cite{gou2018systematic}. Hetero-ReID should face more challenges beyond intra-modality discrepancies, such as HR-to-LR discrepancy, RGB-to-IR discrepancy, Photo-to-Sketch discrepancy, and Image-to-Text discrepancy. These extremely large discrepancies are generated from different modalities. Hence, the approaches of Hetero-ReID should pay more attention to remove the inter-modality discrepancy. Note that we don't attribute the RGB-depth based ReID to our Hetero-ReID, although there are a couple of researches investigated this kind of task~\cite{barbosa2012re,mogelmose2013tri,haque2016recurrent,wu2017robust,munaro20143d}. In realistic applications, the depth channel works as a complement for the RGB channels, and a depth image of a person will not be used to retrieval in an RGB image gallery.

We make a comparison between Homo-ReID and Hetero-ReID, as \tabref{difference} shows. For the media type, Homo-ReID only exploits desired RGB images, while Hetero-ReID takes additional LR/IR/Sketch images or Text description into account. For the participant, Homo-ReID only employs the resources from machine intelligence, while Hetero-ReID also brings in the input from human intelligence, such as sketch images drawn by painters or text descriptions of the suspect by witnesses. For the main challenge, Homo-ReID only needs to deal with the intra-modality discrepancy, such as viewpoint variations, image misaligned or occlusion and illumination changes, while Hetero-ReID should bridge the gaps deriving from both intra- and inter-modality discrepancies. Since additional inter-modality discrepancies are larger than intra-modality discrepancies, Hetero-ReID is more challenging. As intra- and inter-modality discrepancies are essentially different, the methods designed for Homo-ReID cannot be directly used in Hetero-ReID.

There is also a big performance gap between Homo-ReID and Hetero-ReID. The state-of-the-art performances (CMC-1) are reported in \tabref{difference}. They are respectively CAR~\cite{zhou2019discriminative} on Market-1501~\cite{zheng2015scalable} for Homo-ReID, RAIN~\cite{chen2019learning} on MLR-VIPER~\cite{jiao2018deep} for Hetero-ReID LR, D$^2$RL~\cite{wang2019learning} on SYSU-MM01~\cite{wu2017rgb} for Hetero-ReID IR, CDAFL~\cite{pang2018cross} on PKU-Sketch~\cite{pang2018cross} for Hetero-ReID Sketch, and A-GANet~\cite{liu2019deep} on CUHK-PEDES~\cite{li2017person} for Hetero-ReID Text. The performances of Hetero-ReID are still not so satisfied as that of Homo-ReID. However, compared with Homo-ReID, there are not so many publications related to Hetero-ReID. Considering the realistic value and research significance, we feel that it is the right time to call for more attention and research efforts to the area of Hetero-ReID.

\section{Available Datasets and Evaluation Metrics}
\label{sec:datasets}

\subsection{Datasets} 

We summarize available datasets for Hetero-ReID in \tabref{dataset}, including the application scenario, the total number of cameras, identities, and samples. To make a comparison, we also list two typical Homo-ReID datasets, including Market-1501 and MSMT17. The Hetero-ReID datasets are categorized into four types, and we also take notice of the data construction manner, \ie, simulated or really collected. \figref{problem} shows some samples for typical datasets. We have the following summaries: (i) Although many Hetero-ReID LR datasets were constructed, most of them are simulated \emph{w.r.t.} from Homo-ReID datasets. (ii) Only two Hetero-ReID IR datasets were released. SYSU-MM01 is an active IR dataset, while RegDB is a passive one. Only one Hetero-ReID Sketch dataset and one Hetero-ReID Text dataset were constructed. (iii) Compared with Homo-ReID, there is still a lot of room to construct more available datasets. For Hetero-ReID LR, it requires to construct practical datasets instead of the simulated ones. For the other three kinds of application scenarios, datasets under different conditions/styles or with larger scales should be constructed.

\subsection{Evaluation Metrics} 

Evaluation metrics are the same for Homo-ReID and Hetero-ReID. When evaluating Hetero-ReID methods, we often choose the Cumulative Matching Characteristics (CMC)~\cite{wang2007shape} curve and the mean average precision (mAP)~\cite{zheng2015scalable} as the evaluation metrics. In practice, when only one ground truth exists in the gallery, or we are caring about the true matches ranking in top positions of the returning list, the CMC curve is an acceptable selection. On the other hand, if multiple true matches exist in the gallery and we would like to retrieval all the results, mAP is a good selection. Since mAP cares more about the ability of retrieval recall, while CMC does not. Hence, CMC and mAP are always used together for Hetero-ReID.

\tabcolsep=0.5pt
\begin{table}[t]
\renewcommand{\arraystretch}{2}
\caption{Released and freely available datasets. `Appl.', `\#Cam.', `\#ID' and `\#Sam.' respectively represent for the application scenario, the total number of cameras, identities and samples.}
\label{tab:dataset}
\centering
\resizebox{\columnwidth}{!}{
\begin{tabular}{l l c c c r r r}\toprule
\tb{No.}~~ & \tb{Dataset} &\tb{Appl.} &\tb{Type} &\tb{\#Cam.} &\tb{\#ID} &\tb{\#Sam.}\\
\midrule
\midrule
& Market-1501~\cite{zheng2015scalable}    &Desired  &Real      &6   &1,501  &32,668\\
\rowcolor{LightCyan}
& MSMT17~\cite{wei2018person}    &Desired  &Real      &15   &4,101  &126,441\\
\midrule
\circledlr{{1}} &  CAVIAR~\cite{cheng2011custom} &LR &Real &2 &50 &1,000\\
\rowcolor{LightCyan}
\circledlr{{2}} & LR-VIPeR~\cite{li2015multi}   &LR &Simulated &2 &632 &1,264\\
\circledlr{{3}} & LR-3DPES~\cite{li2015multi}   &LR &Simulated &8 &192 &1,011\\
\rowcolor{LightCyan}
\circledlr{{4}} & LR-i-LIDS~\cite{jing2015super}&LR &Simulated &2 &119 &238\\
\circledlr{{5}} & LR-PRID~\cite{jing2015super}  &LR &Simulated &2 &100 &200\\
\rowcolor{LightCyan}
\circledlr{{6}} & SALR-VIPeR~\cite{wang2016scale}&LR &Simulated &2 &632 &1,264\\
\circledlr{{7}} & SALR-PRID~\cite{wang2016scale} &LR &Simulated &2 &450 &900\\
\rowcolor{LightCyan}
\circledlr{{8}} & MLR-VIPeR~\cite{jiao2018deep}&LR &Simulated &2 &632 &1,264\\
\circledlr{{9}} & MLR-SYSU~\cite{jiao2018deep}&LR &Simulated &2 &502 &3,012\\
\rowcolor{LightCyan}
\circledlr{{10}} & MLR-CUHK03~\cite{jiao2018deep}&LR &Simulated &2 &1,467 &14,000\\
\circledir{{11}} & SYSU-MM01~\cite{wu2017rgb} &IR &Real &6 &491 &38,271\\
\rowcolor{LightCyan}
\circledir{{12}} & RegDB~\cite{nguyen2017person}&IR &Real &2 &412 &8,240\\
\circledsk{{13}} & PKU-Sketch~\cite{pang2018cross} &Sketch &Real &2 &200 &400\\
\rowcolor{LightCyan}
\circledtx{{14}} & CUHK-PEDES~\cite{li2017person}&Text &Real &-- &13,003 &80,412\\
\bottomrule
\end{tabular}}
\end{table}

\begin{figure}[t]
    \centering
    \includegraphics[width=\columnwidth]{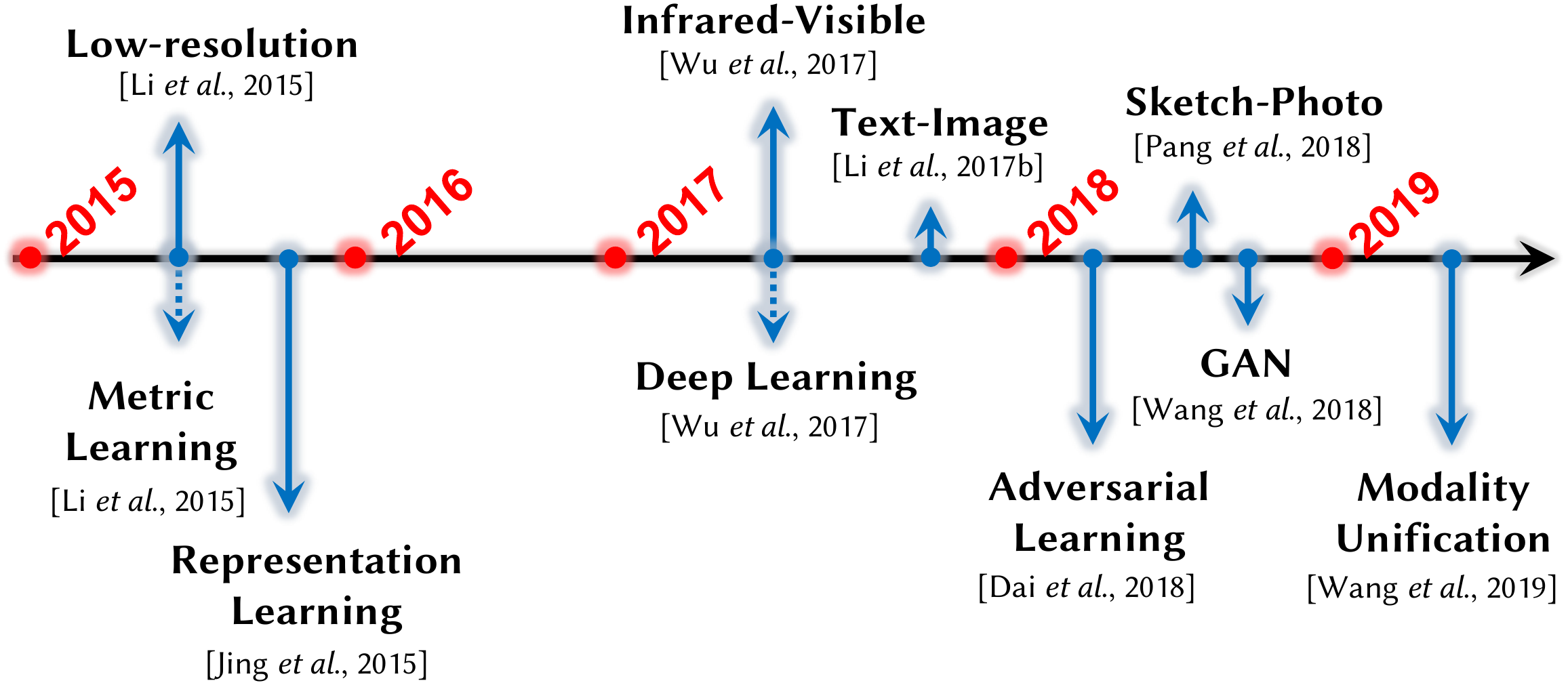}
    \caption{Milestones of existing Hetero-ReID studies. As shown in the figure, Hetero-ReID studies started in 2015. Application scenarios, pipelines and new technical trends are highlighted.}
    \label{fig:milestone}
\end{figure}

\section{Hetero-ReID Methods}
\label{sec:method}
In this section, we first provide the milestones of existing Hetero-ReID studies, and then make a survey of Hetero-ReID methods from two perspectives. From the perspective of application scenario, we classify all related methods into four kinds of Hetero-ReID application scenarios. From the perspective of learning pipeline, we categorize typical methods into three types.

\subsection{Milestones of Existing Hetero-ReID Studies}
Through the unremitting efforts of AI researchers, Hetero-ReID has achieved remarkable success in different aspects. We draw a timeline to introduce important milestones for Hetero-ReID and present them in \figref{milestone}. In particular, `Low-resolution', `Infrared-Visible', `Text-Image', and `Sketch-Photo' indicate milestones of application scenarios. Their typical methods are introduced in \ref{subsec:app_scenario}. `Metric Learning', `Representation Learning', and `Modality Unification' indicate milestones of learning pipelines, which are discussed in \ref{subsec:discussion}. Besides, milestones of new technical trends, such as deep learning, adversarial learning, and GAN, are pointed out as well.

\subsection{From the Perspective of Application Scenario}
\label{subsec:app_scenario}

\heading{Hetero-ReID LR.} Commonly, the resolution of person image changes a lot, due to the variations in the person-camera distance and camera deployment settings. Hetero-ReID LR application scenario attempts to compare images with different resolutions, where low-resolution leads to an extreme loss of appearance information and a large increase of discrepancy. \cite{jing2015super} was the first to investigate the Hetero-ReID LR task. They designed a semi-coupled low-rank discriminant dictionary learning method, constructing a relationship mapping from the features of normal person images to that of LR person images. \cite{li2015multi} proposed a joint multi-scale learning framework by learning metrics on feature domains of two different image scales simultaneously. \cite{wang2016scale} changed the problem to be low-resolution with different scales. They observed that scale-distance functions could be classified, and then learned a surface in function space to separate functions to identify persons. Jiao~\etal~\cite{jiao2018deep} developed a super-resolution and identity joint learning method to improve the Hetero-ReID LR performance. Similar to \cite{jing2015super}, \cite{li2018discriminative} designed a semi-coupled projective dictionary learning model to  bridge the gap across different resolutions. \cite{chen2019learning} combined a resolution adaptation network and a re-identification network together to solve Hetero-ReID LR problem. \cite{wang2018cascaded} first cascaded multiple SR-GANs in series to promote the ability of scale-adaptive super-resolution, then plugged-in a re-identification network to enhance the ability of person representation. \cite{mao2019resolution} jointly trained a Foreground-Focus Super-Resolution module and a Resolution-Invariant Feature Extractor, and then obtained a strong resolution invariant representation. \cite{cheng2020inter} discovered the underlying association knowledge between image SR and ReID, and leveraged it as an extra learning constraint for enhancing the compatibility of SR and ReID models.

\begin{figure*}[t]
\centering
    \includegraphics[width=\textwidth]{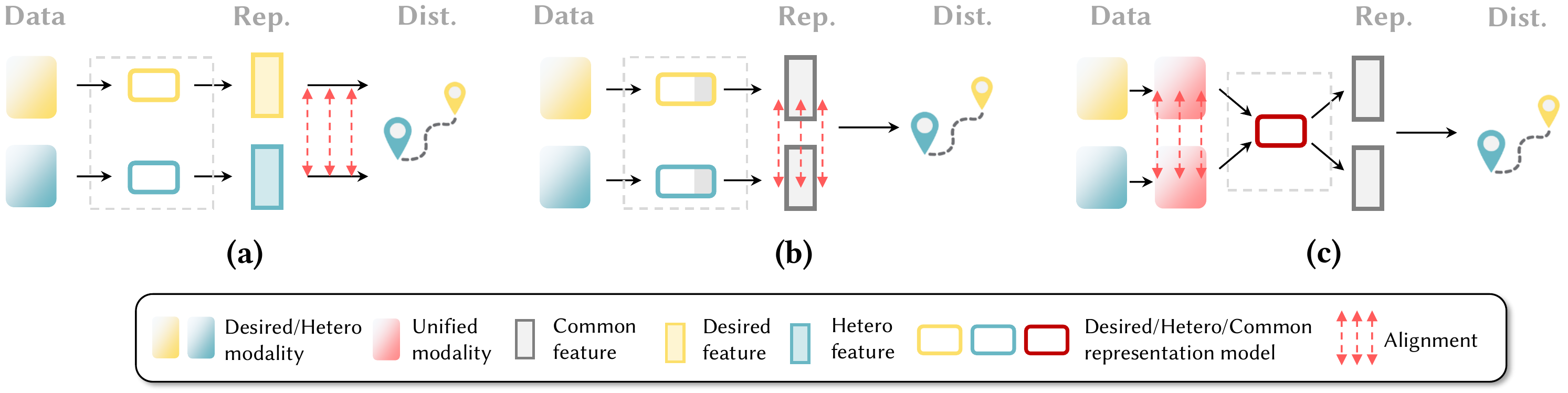}
  \caption{Summary of three learning pipelines. (a) utilizes the metric learning method to learn how to match representations from separate representation models, \ie, `ML'; (b) focuses on learning the common representation of different modalities, \ie, `RL'; (c) pays attention to generating the modality-unified samples and then learning the common representation, \ie, `MU'. In the figure, `D modality' stands for the samples from the desired modality. `H modality' stands for the samples from the heterogeneous modality.}
  \label{fig:framework}
\end{figure*}

\tabcolsep=1pt
\begin{table}[t]
\renewcommand{\arraystretch}{2}
\caption{Representative methods employed in Hetero-ReID. In the datasets column, the number denotes for the certain dataset in \tabref{dataset}.}
\label{tab:comp}
\centering
\resizebox{\columnwidth}{!}{
\begin{tabular}{ l c c l l c }\toprule
\tb{Reference} &\tb{Conf.} &\tb{Appl.}  &\tb{Techn.}   &\tb{Datasets} &{\tb{Pipeline}}\\
\midrule
\midrule
\cite{wang2016scale}    &IJCAI &LR &Subspace Learning
                        &\circledlr{{1}}~\circledlr{{6}}~\circledlr{{7}}
                                                & RL \\
\rowcolor{LightCyan}
\cite{li2015multi}      &ICCV &LR  &Metric Learning
                        &\circledlr{{1}}~\circledlr{{2}}~\circledlr{{3}}
                                                & ML \\
\cite{jing2015super}    &CVPR &LR  &Dictionary Learning
                        &\circledlr{{2}}~\circledlr{{4}}~\circledlr{{5}}
                                                & RL \\
\rowcolor{LightCyan}
\cite{li2018discriminative}&AAAI &LR &Dictionary Learning
                        &\circledlr{{2}}~\circledlr{{4}}
                                                & RL \\
\cite{jiao2018deep}     &AAAI &LR &Super Resolution
                        &\circledlr{{1}}~\circledlr{{8}}~\circledlr{{9}}~\circledlr{{10}}
                                                &  MU\\
\rowcolor{LightCyan}
\cite{wang2018cascaded} &IJCAI &LR &Super Resolution
                        &\circledlr{{1}}~\circledlr{{6}}~\circledlr{{7}}
                                                &  MU\\
\cite{mao2019resolution} &IJCAI &LR &Super Resolution
                        &\circledlr{{1}}~\circledlr{{8}}~\circledlr{{10}}
                                                &  MU\\
\rowcolor{LightCyan}
\cite{chen2019learning} &AAAI &LR &Resolution Adaptation
                        &\circledlr{{1}}~\circledlr{{8}}~\circledlr{{10}}
                                                &  MU\\

\midrule

\cite{ye2018hierarchical}& AAAI &IR &Metric Learning
                        & \circledir{{12}}
                                                & ML \\
\rowcolor{LightCyan}
\cite{wu2017rgb}        & ICCV &IR &Deep Zero-Padding
                        & \circledir{{11}}
                                                & RL \\

\cite{ye2018visible}    & IJCAI &IR &Feature Learning
                        &\circledir{{11}}~\circledir{{12}}
                                                & RL \\
\rowcolor{LightCyan}
\cite{dai2018cross}     & IJCAI &IR  &Feature Embedding
                        & \circledir{{11}}
                                                & RL \\

\cite{wang2019learning} & CVPR &IR &Image Generation
                        & \circledir{{11}}~\circledir{{12}}
                                                & MU\\

\midrule
\rowcolor{LightCyan}
\cite{pang2018cross}    &MM & Sketch &Feature Learning
                        &\circledsk{{13}}
                                                & RL \\

\midrule

\cite{li2017person}     &CVPR &Text &Affinity Learning
                        &\circledtx{{14}}
                                                & ML \\
\rowcolor{LightCyan}
\cite{li2017identity}   &ICCV &Text &Feature Learning
                        &\circledtx{{14}}
                                                & RL \\

\cite{chen2018improving2}  &WACV &Text & Patch Matching
                        &\circledtx{{14}}
                                                & ML \\
\rowcolor{LightCyan}
\cite{chen2018improving}  & ECCV &Text &Association Learning
                        &\circledtx{{14}}
                                                & RL \\

\cite{liu2019deep}  & MM &Text &Adversarial Learning
                        & \circledtx{{14}}
                                                & RL \\
\bottomrule
\end{tabular}}
\end{table}

\heading{Hetero-ReID IR.} The visible cameras are not able to capture clear and valid appearance information under poor illumination environments (\eg~during the nighttime), which limits the applicability of ReID in practical surveillance applications. Hetero-ReID IR application scenario provides a good supplement for nighttime surveillance applications, which also introduces a large intra-modality discrepancy. \cite{wu2017rgb} was the first to investigate Hetero-ReID task in the Hetero-ReID IR task. They proposed a deep zero-padding module to improve the one-stream network, making the implicit network structure more suitable. \cite{ye2018hierarchical} jointly optimized the modality-specific and modality-shared metrics and designed a hierarchical cross-modality matching model. \cite{ye2018visible,ye2019bi} attempted to learn discriminative features with a bi-directional dual-constrained top-ranking loss function. \cite{dai2018cross} proposed to learn discriminative common representations with a generator for learning image representations and a discriminator for discriminating the modalities of RGB and IR images. \cite{wang2019learning} presented a new framework, taking advantage of CycleGAN to reduce modality discrepancy from image-level and advantage of sophisticated Homo-ReID models to reduce appearance discrepancy from feature-level. \cite{wang2020cross} jointly exploited pixel alignment and feature alignment by generators. \cite{li2020infrared} introduced an auxiliary X modality as an assistant and an X-Infrared-Visible three-mode learning framework. \cite{kansal2020sdl} designed a network with disentanglement loss to distill identity features and dispel spectrum features. \cite{choi2019hi} also attempted to disentangle ID-discriminative factors and ID-excluded factors from images, and they designed an ID-preserving person image generation network to implement the idea. \cite{jin2020style} also proposed to distill identity-relevant feature from the removed spectrum style information and restitute it to the network to ensure high discrimination. \cite{lu2020cross} explored the modality-shared information and the modality-specific characteristics to boost the performance. \cite{yang2020mining} proposed a bi-directional random walk scheme to mining more reliable relationships between images by traversing heterogeneous manifolds in the feature space of each modality. Note that the method \cite{yang2020mining} performs as a re-ranking method thus obtains relative high performances.

\heading{Hetero-ReID Sketch.} When someone reports a lost child case while the child was not identified by the surveillance camera, we call for an algorithm to automatically pick out the potential candidates according to a sketch by an artist's drawing. Consequently, investigators will be able to narrow down and hence expedite their search. \cite{pang2018cross} was the first and only one to investigate Hetero-ReID task in the Sketch application scenario. They not only raised a dataset, but also proposed to learn identity features and modality-invariant features by exploiting a cross-modal adversarial feature learning framework.

\heading{Hetero-ReID Text.} Searching a person with free-form natural language descriptions can be widely applied. In the early years, some researches started to investigate the ReID with attribute-based queries~\cite{feris2014attribute,ye2015specific,wang2015multi,yin2018adversarial}. Hetero-ReID Text application scenario denotes a situation that the queries are natural language descriptions. \cite{li2017person} evaluated and compared a wide range of possible models, and proposed an RNN with Gated Neural Attention mechanism. \cite{li2017identity} learned to embed cross-modal features, and then refined the matching results with a co-attention mechanism. \cite{zhang2018deep} designed a cross-modal projection matching loss and a cross-modal projection classification loss for learning discriminative image-text embeddings. \cite{chen2018improving2} proposed a patch-word matching model and designed an adaptive threshold mechanism into the model. \cite{chen2018improving} attempted to build global and local image-language associations, which enforce semantic consistencies between local visual and linguistic features. \cite{liu2019deep} proposed a deep adversarial graph attention convolution network.

\heading{Summary.} From the perspective of application scenario, we make the following summaries:
\begin{itemize}
\item \emph{Most of the methods selected a deep learning framework}. It is probably because all the heterogeneous application scenarios are raised in recent years, and deep learning methods are in their high-speed development period. In general, deep learning methods are superior in shared feature learning and classification tasks when a significant amount of training samples are available.
\item \emph{Different methods have different focuses}. To fill the gaps between the desired modality and other heterogeneous modalities, some methods (\eg~\cite{li2015multi}) try to learn a metric, others (like \cite{ye2018hierarchical}) attempt to learn shared features, and there are also methods (for example \cite{wang2019learning}) unifying the modality on the data level. In \ref{subsec:discussion}, we will make an analysis and discussion.
\item \emph{The existing researches in each application scenario still have many limitations}. For the Hetero-ReID LR application scenario, recent researches are mainly evaluated on the simulated datasets. For the Hetero-ReID Sketch application scenario, only one research touched this direction, due to the hardness of building a benchmark. 
\end{itemize}

\subsection{From the Perspective of Learning Pipeline}
\label{subsec:discussion}
Representative methods employed in Hetero-ReID are compared in \tabref{comp}. We list some key and valuable information, including the reference paper, conference (`Conf.'), application scenario (`Appl.'), technology (`Techn.'), used datasets performance (`CMC-1') and Learning pipeline. We can find that all representative works are published in the recent five years. It means that Hetero-ReID is a relatively new research topic. Among all the application scenarios, Hetero-ReID Sketch is less investigated. As described in the sections above, we consider that if more datasets are available, more works could be conducted and published. Particularly, to address the inter-modality discrepancy, in our opinion, representative methods can be categorized into three kinds of learning pipelines.

\figref{framework} shows the diagrams of the learning pipelines. Pipeline (a) employs the metric learning method to learn how to match the features from separate representation learning models, and the representation learning models are trained separately with single modality data samples (represented as `ML'). Pipeline (b) focuses on learning shared feature models of different modalities, and training data come from both modalities (represented as `RL'). Pipeline (c) pays attention to generating the unified-modality samples (represented as `MU'), for example, using a super-resolution method to generate high-resolution images from low-resolution images, or using some image generation method to generate infrared images from RGB images and RGB images from infrared images.

As shown in \tabref{comp}, for each kind of application scenario, from top to bottom, the methods become more effective. Generally, pipeline `MU' performs better than pipeline `RL', and better than pipeline `ML'. Hence, unifying the modalities of samples is an effective way to fill the modality gap. For the Hetero-ReID LR application scenario, the super-resolution methods can be used, such as \cite{wang2018cascaded,mao2019resolution}. Thanks to the developments of GAN, unifying the modalities is considerable as \cite{wang2019learning}.

\section{Conclusion and Future Directions}
\label{sec:conc}
This paper overviews recent developments of Hetero-ReID. It covers most of the literature on Hetero-ReID. We summarized available datasets, the widely employed methods, and compared the existing techniques. We believe that Hetero-ReID will continue to be an active and promising research area with broad potential applications. Many issues in Hetero-ReID, however, still remain.

\begin{figure}[t]
\centering
    \includegraphics[width=.75\columnwidth]{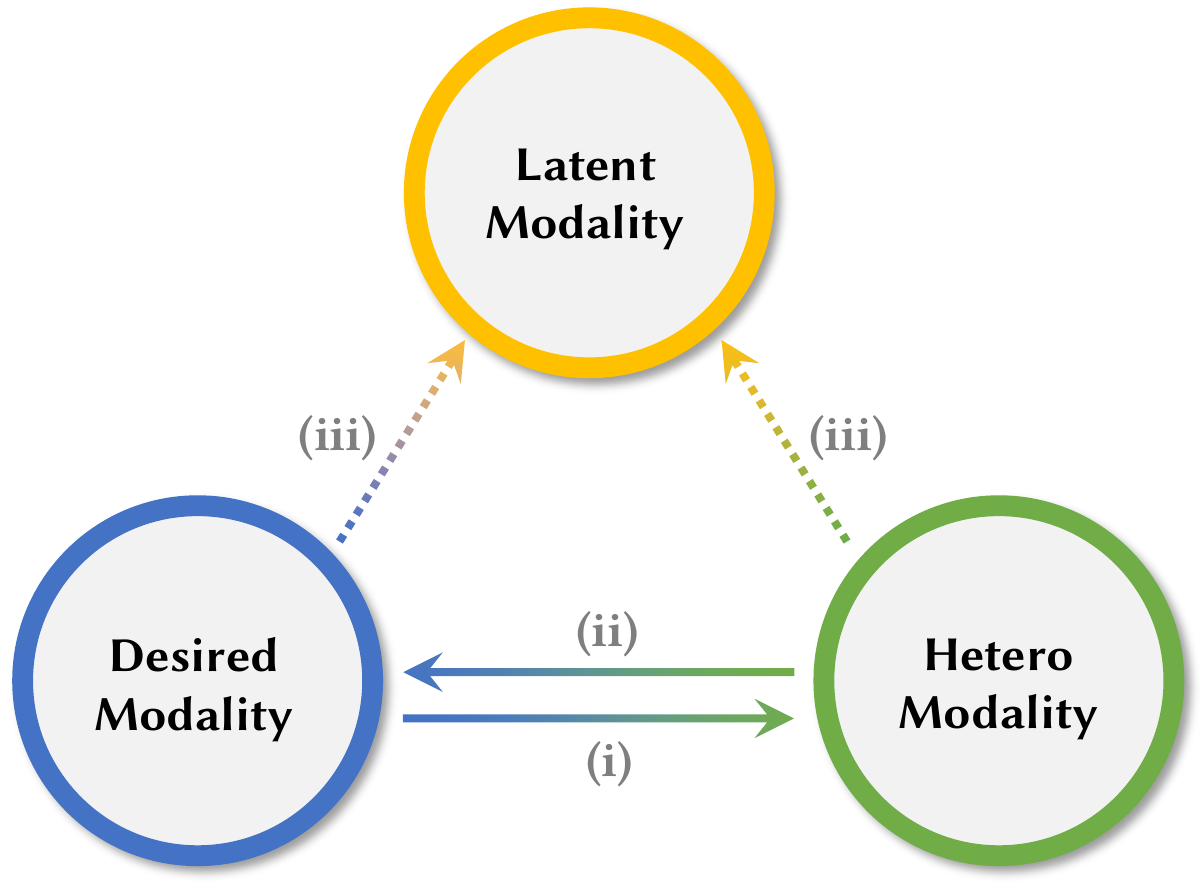}
  \caption{Three kinds of modality transfer. To unify the modality, we can transfer (i) from heterogeneous modality to desired modality, (ii) from desired modality to heterogeneous modality, and (iii) from both desired and heterogeneous modalities to a latent modality.}
  \label{fig:relation}
\end{figure}

\begin{itemize}
\item \emph{Dataset Construction}. For Hetero-ReID LR application scenario, the researchers can only get simulated datasets. For the other types of applications, although the researchers can achieve practical datasets, the choice is limited. In addition, large scale Hetero-ReID datasets are also required. It is somehow urgent for us to built new datasets, and push forward this research area. On the other hand, synthesizing more realistic datasets (like \cite{zeng2019illumination}) is also a good way to make one step forward.
\item \emph{Taking Advantages of Homo-ReID Datasets and Methods}. There are sufficient datasets and methods in Homo-ReID. It is easy to address the intra-modality discrepancy in the desired modality. Obviously, the feature space of the desired modality can be well-constructed. Since the heterogeneous modality samples are not enough, it is hard to build a perfect heterogeneous feature space. However, it is reasonable to project the Hetero-ReID samples into the desired modality (like \cite{yang2020mining}).
\item \emph{Human Interaction and Crowd-sourcing}. For Hetero-ReID Sketch and Text application scenarios, human intelligence joins into the process of ReID. Human intelligence is sometimes subjective and incomplete. So we should consider how to mine and integrate useful information to help search the target in the surveillance system. On the other hand, a lot of witnesses will provide their cues. Each person may have a different view so that the crowd-sourcing cues may be diverse to each other. We should design a strategy to remove conflict and filter valuable information.
\item \emph{Investigation on Unifying the Modality}. For Hetero-ReID LR application scenario, some methods~\cite{wang2018cascaded,mao2019resolution} attempted to use super-resolution technologies to unify the modality, where they transfer the heterogeneous modality LR image to the desired modality. For Hetero-ReID IR application scenario, with the help of CycleGAN, one work~\cite{wang2019learning} not only transferred the image from heterogeneous (IR) modality to desired (RGB) modality but also transferred the image from desired modality to heterogeneous modality. However, for the Hetero-ReID Text and Sketch application scenarios, no work has investigated to unify the data modality. On the other hand, we can also investigate to unify the modality to a latent modality as \figref{relation} shows, for example, a middle-level resolution for the Hetero-ReID LR application scenario and a hyperspectral image for the Hetero-ReID IR application scenario.
\item \emph{Integrating Multiple Hetero-ReID Application Scenarios}. For a practical system, it may require different kinds of inputs to search out the target. The Hetero-ReID application scenarios can be equipped in different stages. If we can integrate different kinds of inputs, more valuable information could be used for retrieval, since different inputs have different attentions and views for the target. It would raise a novel multiple cross-modality research task.
\item \emph{Considering the Privacy Issue}. It is fine to conduct academic research on public datasets. But with the leakage of personal image data, privacy concerns are raising nowadays when algorithms need to be applied to practical applications. Recently, pioneer works~\cite{mirjalili2017soft} have explored to hide some of private information presented in the images. Further research on the principles of visual cryptography, signal mixing and image perturbation to protect users’ privacy on person templates are essential for addressing public concern on privacy. In particular, for Hetero-ReID, since it contains multi-modality data, person templates of different modalities or a latent modality should be investigated to protect.

\end{itemize}

\section*{Acknowledgement}
\small{This work was supported partly by the JST CREST Grant JPMJCR1686, partly by the Grant-in-Aid for JSPS Fellows 18F18378, and partly by the Microsoft Collaborative Research Grant.}

\balance
{
\bibliographystyle{named}
\small
\bibliography{ijcai20}
}

\end{document}